# Multi-Document Summarization via Discriminative Summary Reranking

Xiaojun Wan[1], Ziqiang Cao[1], Furu Wei[2], Sujian Li[1] and Ming Zhou[2]

[1]Peking University
[2]Microsoft Research Asia
[1]{wanxiaojun, ziqiangyeah, lisujian}@pku.edu.cn
[2]{fuwei, mingzhou}@microsoft.com

**Abstract**

Existing multi-document summarization systems usually rely on a specific summarization model (i.e., a summarization method with a specific parameter setting) to extract summaries for different document sets with different topics. However, according to our quantitative analysis, none of the existing summarization models can always produce high-quality summaries for different document sets, and even a summarization model with good overall performance may produce low-quality summaries for some document sets. On the contrary, a baseline summarization model may produce high-quality summaries for some document sets. Based on the above observations, we treat the summaries produced by different summarization models as candidate summaries, and then explore discriminative reranking techniques to identify high-quality summaries from the candidates for difference document sets. We propose to extract a set of candidate summaries for each document set based on an ILP framework, and then leverage Ranking SVM for summary reranking. Various useful features have been developed for the reranking process, including word-level features, sentence-level features and summary-level features. Evaluation results on the benchmark DUC datasets validate the efficacy and robustness of our proposed approach.

## Introduction

Given a set of documents about a topic, multi-document summarization systems aim to produce a short and fluent summary to deliver the salient information in the document set. Most existing summarization systems are based on sentence extraction, and they rely on a specific method to rank some kinds of units (e.g. words, bigrams, or sentences) and then extract summary sentences according to the ranking results. With the development of the DUC and TAC benchmark tests, document summarization has been well studied and many different summarization methods have been proposed, e.g., centroid-based method (Radev et al. 2004), graph-based ranking methods (Erkan and Radev 2004) and ILP-based methods (McDonald 2007; Gillick et al. 2008).

In an existing document summarization system, a single summarization model (i.e., a summarization method with a specific parameter setting) is used for extracting summaries from different document sets. For example, there are 50 different document sets in the DUC2004 dataset, and a summarization system usually adopts a single summarization model (e.g. an ILP-based method with a specific parameter setting) to extract summaries for all the 50 document sets. The common assumption is that a single summarization model can well deal with all the different document sets. However, according to our quantitative data analysis, none of the existing summarization models can always produce high-quality summaries for different document sets, and even a summarization model with good overall performance may produce low-quality summaries for some document sets. On the contrary, a baseline summarization model may produce high-quality summaries for some document sets. The reason lies in the different characteristics of different document sets and none of the summarization models can be fit for all different document sets.

Based on the above observations, we attempt to improve the overall summarization performance by leveraging the summaries produced by multiple different summarization models (i.e., different summarization methods, or a summarization method with different parameter settings) for each document set. We treat the summaries produced by different summarization models as candidate summaries, and then explore discriminative reranking techniques to identify high-quality summaries from the candidates for difference document sets. In this way, we can take advantage of different summarization models and produce better summaries for different document sets, and thus achieve better overall performance. In particular, we propose to extract a set of candidate summaries for each document set based on an ILP framework. In order to discriminate high-quality summaries from low-quality summaries, we adopt Ranking SVM and develop various useful features for the reranking process, including word-level features, sentence-level features and summary-level features. Evaluation results on the benchmark DUC datasets show that our proposed approach can outperform each single summarization model used and achieve state-of-the-art performances in terms of ROUGE scores. The efficacy of each group of features is also verified.

To the best of our knowledge, we are the first to apply discriminative reranking techniques for extractive document summarization.

## Related Work

Multi-document summarization methods can be extraction-based or abstraction-based, and we focus on extractive summarization methods in this paper. Extractive summarization methods usually produce a summary by selecting some original sentences in the document set. Sentences can be scored by employing rule based methods to simply combine a few feature weights, e.g., the centroid-based method (Radev et al. 2004) and NeATS (Lin and Hovy 2002). Machine learning techniques have been used for better combining various sentence features (Ouyang et al. 2007; Shen et al. 2007; Schilder and Kondadadi 2008; Wong et al. 2008). Many advanced methods have been proposed for extractive summarization in recent years, which are based on various techniques: budgeted median method (Takamura and Okumura 2009), A* search algorithm (Aker et al. 2010), minimum dominating set (Shen and Li 2010), matrix factorization (Wang et al. 2008), topic model (Wang et al. 2009) Integer Linear Programming (McDonald 2007; Gillick et al. 2008; Gillick and Favre 2009; Li et al. 2013), and submodular function (Lin and Bilmes 2010; Li et al. 2012). Graph-based methods have also been proposed for various summarization tasks, such as LexRank (Erkan and Radev 2004), TextRank (Mihalcea and Tarau 2005) and ClusterCMRW (Wan and Yang 2008). Furthermore, ensemble methods have also been used for sentence ranking. For example, Wang and Li (2010) propose a weighted consensus method to aggregate different sentence ranking results by different summarization methods.

However, all the above studies focus on sentence scoring and ranking, and none of them has attempted to rank summaries directly. Different from previous studies, we rank summaries directly in this study. The advantage of ranking summaries is that we can optimize the summarization performance directly based on the characteristics of the summaries.

Automatic summary evaluation is partially related to this work. Most researches in this area focus on how to measure the quality (i.e., content and readability) of a summary when one or more reference summaries written by human experts are given (Lin and Hovy 2003; Hovy et al. 2006; Pitler et al. 2010; Lin et al. 2012). In particular, ROUGE is one of the most popular metrics for comparing peer summaries with reference summaries. In recent years, several pilot studies have investigated to automatically assess the qualities of peer summaries without reference summaries (Saggion et al. 2010; Louis and Nenkova 2013), and they mainly rely on the similarity (e.g. JS divergence and KL divergence) between the peer summary and the source document text. However, only the summary-to-document similarity is not adequate for ranking summaries, as shown in our experiments, and we have to develop more useful features on different levels.

## Data Analysis and Motivation

In this section, we take the DUC2004 dataset as example and perform a quantitative analysis on the dataset to validate our assumption and present our motivation.

In the multi-document summarization task (i.e. task 2) on DUC2004, there are a total of 50 English document clusters and each cluster contains 10 news document on average. Given each document cluster, the task aims to create a short summary (<= 665 bytes) of the cluster. Reference summaries for each cluster have been created for evaluation. There were a total of 34 runs submitted by 15 teams, and these runs were produced by different summarization models (i.e., different summarization methods with different parameter settings). Therefore, all the runs can represent a variety of different summarization models. The overall performances of these summarization models range from 1.851 to 9.178 in terms of ROUGE-2 recall (%)[1].

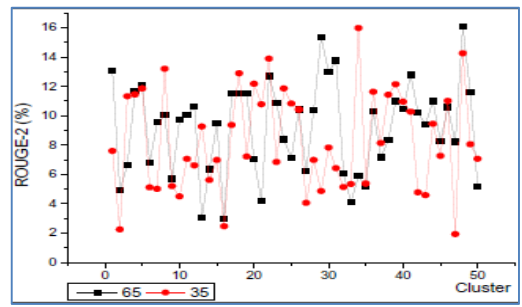

---

[1] Note that the ROUGE scores reported in this paper are the percentage values, which means the real ROUGE scores times 100. We used ROUGE-1.5.5 toolkit for evaluation in this paper.

Figure 1. ROUGE-2(%) comparison of sample runs on different document clusters.

Since the overall performance of a summarization model is the average of the performance values of the model across the 50 document clusters, we now compare the detailed performance values of two different runs (i.e. 65 and 35) on different document clusters, as shown in Figure 1. We can see from the figures that different runs produced by different summarization models have variable performances on different document clusters. For example, run 65 has the best overall performance, but it achieves the best scores over only 7 document clusters (e.g., clusters 1, 29, 47); moreover, its performance values over a few document clusters (e.g., clusters 13, 21, 34) are very low. On the other hand, run 35 has a lower overall performance and ranks 6th, but it can achieve the highest scores over 4 document clusters (e.g. clusters 3, 20, 21, 34), and it can produce better summaries than run 65 for 19 clusters.

Ideally, if we can select the best one from the summaries of different runs for each document cluster, we can obtain the upper bound performance, as shown in Table 1. The average performance of all runs are also presented. We can see that the upper bound performance is much higher than the top performance of the submitted runs.

Table 1: The upper bound vs. top run

|  | ROUGE-2(%) |
|---|---|
| Upper Bound | 11.76 |
| Average | 7.010 |
| Top run (65) | 9.178 |
| Worst run (111) | 1.851 |

To say the least, if we can select the one close to the best for each document cluster, we can still achieve high overall performance. So the question is whether and how we can select the best one or the one close to the best from a set of candidate summaries for each document cluster. Fortunately, this is a typical ranking problem and we can leverage learning to ranking techniques to address it, as described in the next section.

## Proposed Approach

Given a document set, our proposed approach aims to select for the document set a high-quality summary from a set of candidate summaries of various qualities. It consists of two stages: the first stage aims to extract candidate summaries for each document set, and the second stage aims to rerank the candidate summaries and select the best one as the final summary for each document set. The technical details of the two stages will be described in the following subsections, respectively.

### Candidate Summary Extraction

The summaries of the submitted runs can be simply used as candidate summaries for the DUC document sets, but in practice, we do not have "submitted runs" for a new document set, so we have to develop a method to produce candidate summaries for any given document set. There are several ways to achieve this goal. For example, we can develop different summarization methods to extract candidate summaries, or we can make use of a single summarization method by using different parameter values to extract summaries, or we can combine the above two ways.

In this study, we propose to extract candidate summaries for a document set based on an ILP framework, which is very popular in the summarization area in recent years. Particularly, we leverage the following ILP formulation for sentence extraction and summary generation:

$$\max\{\lambda \sum_{i \in S} \frac{l_i}{L} u_i x_i + (1-\lambda) \sum_{j \in C} \frac{1}{L} v_j y_j\} \quad (1)$$

s.t.
$$\sum_i l_i x_i \leq L \quad (2)$$
$$x_i O_{i,j} \leq y_j, \quad \forall i,j \quad (3)$$
$$\sum_i x_i O_{i,j} \geq y_j, \quad \forall j \quad (4)$$
$$x_i \in \{0,1\}, \quad \forall i$$
$$y_j \in \{0,1\}, \quad \forall j$$

where the notations are defined as follows:
$S$: the set of sentences in a document set;
$C$: the set of words in a document set;
$u_i$: the importance score of sentence $i$, which is learned by a regression method described later;
$v_j$: the importance score of word $j$, which is learned by a regression method described later;
$O_{i,j}$: the indicator of whether word $j$ occurs in sentence $i$;
$x_i$: the indicator of whether sentence $i$ is selected into the summary;
$y_j$: the indicator of whether word $j$ appears in the summary;
$l_i$: the length of sentence $i$;
$L$: the length limit of the summary;
$\lambda$: the parameter between [0, 1] to control the influences of two parts;

In the above formulation, both the scores of sentences and words are considered. Constraint (2) ensures the summary's length limit. Constraints (3) and (4) ensure that a sentence is selected iff the words in the sentence are selected, and a word is selected if at least one sentence containing the word is selected. The first part aims to select sentences with higher importance scores. We add the sentence length ratio $\frac{l_i}{L}$ as a multiplication factor in order to penalize the very short sentences, or the objective function tends to select more and shorter sentences. At the same time, the objective function does not tend to select the very long sentences. The total length of the sentences selected is fixed. So if the objective function tends to select the longer sentences, the fewer sentences can be selected. A tradeoff needs to be made between the number and the average length of the sentences selected. The second part aims to let the summary contain important words as many as possible. This part can address the redundancy issue of the summary. The intuition is that the more unique words the summary contains, the less redundancy the summary has.

In order to better assess the importance of each sentence and word, we leverage the support vector regression (SVR) method implemented in LIBLINEAR [2] to learn the importance scores $u_i$ and $v_j$. In the training phase, the frequency of a word in the reference summaries is used as the target score of the word, and the maximum similarity between a sentence and the sentences in the reference summaries is used as the target score of the sentence. There are a total of 15 features used for word score regression, including term frequency (TF), document frequency (DF), POS-based features (whether a word is a noun/verb/adjective/adverb), NER-based feature (whether a word belongs to a named entity) and number-based feature (whether a word is a number), and also the features extracted from the sentences containing the word (e.g. max/min positions, etc.) There are a total of 13 features used for sentence score regression, including sentence's position, length, number of subsentences, depth of the parse tree, proportion of stopwords, mean TF of words, mean DF of words, proportion of words in Noun/Verb/Adjective/Adverb/NER/number.

Based on the above ILP formulation, we first change $\lambda$ from 0 to 0.9 with a step of 0.1, and thus produce 10 candidate summaries for each document set. Moreover, for each value of $\lambda$, we iteratively produce 10-best summaries by adding the following new constraint at each iteration:

$$\frac{\sum_{i \in X_k} x_i}{|X_k|} \leq \beta, \ \forall k \qquad (5)$$

where $X_k$ is the set of all sentences in all the summaries produced from the first to the $k$-th iteration. $\beta$ controls how the new summary at the $(k+1)$-th iteration can resemble existing summaries. We simply set $\beta$ to 0.6 in our experiments, which means at most 60% sentences in the summary produced at the $(k+1)$-th iteration are allowed be the same with the sentences in all the previous summaries. For example, if we have selected sentences $X_1$={1, 2, 3, 4, 5} at the first iteration, we are allowed to select at most 60% the same sentences from $X_1$ at the second iteration, for example, {1, 2, 3, 6, 7}, and then we are allowed to select at most 60% the same sentence from $X_2$={1, 2, 3, 4, 5, 6, 7} at the third iteration, and so on.

Finally, we can produce a total of 100 summaries for each document set, and these summaries are treated as candidate summaries.

## Summary Reranking

**Method**

After we obtain a set of candidate summaries for a document set, we adopt the learning to rank techniques to discriminate the high-quality ones from the low-quality ones. In recent years, various learning to ranking algorithms have been proposed in the information retrieval field, such as RankSVM (Joachims 2002), RankNet (Burges et al. 2005), RankBoost (Freund et al. 2003), etc. Without loss of generality, we adopt RankSVM for summary reranking, because RankSVM is the most popular learning to rank technique and it has been successfully used in many applications. Due to page limit, the comparison of different learning to ranking algorithms is out of the scope of this paper.

RankSVM (Ranking SVM) is a pairwise approach for learning to rank. It makes use of the regular SVM QP optimization and trains for a classification of order of pairs. In this study, we use the $SVM^{rank}$ toolkit [3] in our experiments.

In the training phase, since we have the reference summaries for each document set, we use the ROUGE-2 recall scores between the candidate summaries and the reference summaries to derive the ranking order of the candidate summaries.

**Features**

Given a document set and a set of candidate summaries, we can extract a variety of features on different levels to indicate different aspects of the quality of each candidate summary. Note that reference summaries cannot be used for feature extraction. Three group of features have been extracted for the reranking process: word-level features, sentence-level features and summary-level features. The details of the three groups of features are presented in Tables 2, 3 and 4, respectively. The values of these features are scaled to [0,1].

Table 2. Word-level features

| Feature | Description |
| --- | --- |
| **TF** | Sum of word frequency in a summary, where word frequency is computed from the document set; |
| **DF** | Sum of document frequency of words in a summary, where document frequency is computed from the document set; |
| **POS** | The proportion of noun/verb/adverb/adjective words in a summary; |
| **NER** | The ratio of named entity number to the summary length; |
| **Stopword** | The ratio of stopword number to the summary length; |
| **Number** | The ratio of number word count to the |

---

[2] http://www.csie.ntu.edu.tw/~cjlin/liblinear/

[3] http://www.cs.cornell.edu/people/tj/svm_light/svm_rank.html

| Feature | Description |
|---|---|
| | summary length; |
| **Unique Words** | The ratio of the number of unique words in a summary to the summary length; |
| **Lead Words** | The ratio of the number of words in a summary which appear in the first sentences of documents to the summary length; |

Table 3. Sentence-level features

| Feature | Description |
|---|---|
| **Sentence Length** | The min/max/mean length of sentences in a summary; |
| **Position** | The mean/max position weight of sentences in a summary, where the position weight is computed as $1 - \frac{(position-1)}{(sentence\ count-1)}$; |
| **Sentence Number** | The number of sentences in a summary; |

Table 4. Summary-level features

| Feature | Description |
|---|---|
| **Sum-Doc Cosine with TFIDF** | The cosine similarity between a summary and the document set, where the summary and the concatenated text for the document set are represented by two TFIDF vector of words; |
| **Sum-Doc JS** | The Jensen Shannon divergence between a summary and the document set; |
| **Sum-Doc Word Overlap** | The word overlap similarity between a summary and the document set; |
| **Sum-Doc Cosine with Embedding** | The cosine similarity between a summary and the document set, where the summary text and the concatenated text for the document set is represented by averaging the embedding vectors of all words in the text[4]; |
| **Sum-Sum Word Overlap** | The average word overlap similarities between a summary and other candidate summaries; |
| **Sum-Sum Cosine with Embedding** | The average cosine similarity between a summary and other candidate summaries, where each summary text is represented by averaging the word embeddings; |

## Evaluation

### Evaluation Setup

In the experiments, we used three benchmark DUC datasets for evaluation: DUC2001, DUC2002 and DUC2004. In each dataset, generic summaries are required to be created for different news clusters (i.e. document set). Reference summaries have been manually provided for each document set. The datasets are summarized in Table 5.

---

[4] Word embeddings are downloaded from http://ml.nec-labs.com/senna/. The dataset contains 130000 words and each word is associated to an embedding with a dimension of 50.

Table 5. Summary of datasets

| | DUC 2001 | DUC 2002 | DUC2004 |
|---|---|---|---|
| **Task** | Task 2 | Task 2 | Task 2 |
| **Number of documents** | 309 | 567 | 500 |
| **Number of clusters** | 30 | 59 | 50 |
| **Data source** | TREC-9 | TREC-9 | TDT |
| **Summary length** | 100 words | 100 words | 665 bytes |

In the experiments, we used DUC2002 and DUC2004 as test set. When DUC2002 was used as test set, DUC2001 and DUC2004 were used as training set. When DUC2004 was used as test set, DUC2001 and DUC2002 were used as training set. Both the regression models and the Ranking SVM model were trained and tuned on the training set. The models were then applied on the test set.

In this study, we used ROUGE-2 recall score (%) as the evaluation metric, because ROUGE-2 was the most reliable evaluation metric for document summarization and it has been shown to be highly correlated with human judges. Due to page limit, other ROUGE scores were ignored in this paper, as the conclusions based on these scores are the same with that based on ROUGE-2.

### Evaluation Results and Discussion

**Main Results**

In our proposed approach, we have 100 summarization models based on the ILP framework to produce a total of 100 different candidate summaries for each document set. We compare our reranking approach with the best model and the worst model among them. We also compute the average scores of the 100 models. The upper bound scores are computed by selecting the best summary for each document set. The comparison results are shown in Table 6. We can see that on both datasets, the performance gap between the best model and the worst model is very big, and the upper bound is much higher than that of the best model. Moreover, our proposed reranking approach can sensibly outperform the best models on both datasets. The results verify the effectiveness of our proposed approach.

In our proposed approach, we use RankSVM to rerank the candidate summaries. As mentioned earlier, Louis and Nenkova (2013) have proposed to use JS divergence and KL divergence between a candidate summary and a document set to automatically evaluate the summary's quality without reference summaries, and therefore, JS divergence and KL divergence can be used for reranking the candidate summaries. The comparison between RankSVM, JS divergence and KL divergence is shown in Table 7. We can see that RankSVM significantly outperforms JS divergence and KL divergence. The results verify the effectiveness of the use of RankSVM with multiple useful features for reranking candidate summaries.

Table 8 further compare our proposed approach with a variety of state-of-the-art methods, besides the best DUC participating system (top run). The ROUGE scores of the methods are directly borrowed from the corresponding literatures. As shown in the table, our proposed approach can achieve state-of-the-art performance.

In Table 9, we compare different feature sets in our proposed approach. "w/o word-level" means removing the word-level features from the feature set. "w/o embedding" means removing the features relying on word embedding. We can see that all kinds of features are beneficial for the reranking process, including the features with word embedding.

Table 6. Comparison results[5]

|  | DUC2002 | DUC2004 |
|---|---|---|
| Upper Bound | 11.199 | 11.85 |
| Average | 7.298 | 8.754 |
| Best Model | 8.192 [7.238-9.138] | 9.760 [8.909-10.617] |
| Worst Model | 6.679 [5.583-7.764] | 7.526 [6.711-8.357] |
| Our Approach | **8.555** [7.616 – 9.520] | **10.051** [9.334-10.680] |

Table 7. Comparison of reranking strategies

|  | DUC2002 | DUC2004 |
|---|---|---|
| RankSVM | **8.555** [7.616 –9.520] | **10.051** [9.334-10.680] |
| JS Divergence | 7.209 [6.404-8.053] | 8.876 [8.083-9.669] |
| KL Divergence | 7.665 [6.779 -8.604] | 8.660 [7.877-9.461] |

Table 8. Comparison with state-of the-art methods

|  | DUC2002 | DUC2004 |
|---|---|---|
| Our Approach | **8.555** | **10.051** |
| ClusterHITS (Wan and Yang 2008) | 8.135 | - |
| ILP (McDonald 2007) | 7.2 | - |
| MSSF (Li et al. 2012) | - | 9.897 |
| BSTM (Wang et al. 2009) | - | 9.01 |
| MDS (Shen and Li 2010) | - | 8.934 |
| Top Run | 7.642 | 9.178 |

Table 9. Feature analysis results

|  | DUC2002 | DUC2004 |
|---|---|---|
| All features | **8.555** | **10.051** |
| w/o word-level | 8.245 ↓ | 9.988 ↓ |
| w/o sentence-level | 8.265 ↓ | 10.06 |
| w/o summary-level | 8.4 ↓ | 9.65 ↓ |
| w/o embedding | 8.458 ↓ | 10.037 ↓ |

**Results with More Candidates**

In the above experiments, we produce 100 candidate summaries for each document set. We now add more candidates for reranking by produce a set of candidate summaries based on the LexRank method. We range the damping factor in the LexRank algorithm from 0.1 to 0.9 with a step of 0.05. The results with more candidates are shown in Table 10. We can see that the upper bound scores have been improved due to the more candidate summaries. Compared with Table 5, the performance scores of our proposed approach have also been improved slightly. The new feature comparison results are presented in Table 11, and all kinds of features are still very useful.

---

[5] The 95% confidence interval for each ROUGE score is reported in brackets. For the upper bound and average scores, the confidence intervals are not available.

Table 10. Comparison results with more candidates

|  | DUC2002 | DUC2004 |
|---|---|---|
| Upper Bound | 11.486 | 12.064 |
| Average | 7.367 | 8.763 |
| Best Model | 8.192 [7.238-9.138] | 9.760 [8.909-10.617] |
| Worst Model | 6.679 [5.583-7.764] | 7.526 [6.711-8.357] |
| Our Approach | **8.649** [7.604-9.682] | **10.154** [9.337-11.074] |

Table 11. Feature analysis results with more candidates

|  | DUC2002 | DUC2004 |
|---|---|---|
| All features | **8.649** | **10.154** |
| w/o word-level | 8.023 ↓ | 10.009 ↓ |
| w/o sentence-level | 8.491 ↓ | 9.893 ↓ |
| w/o summary-level | 8.457 ↓ | 10.036 ↓ |
| w/o embedding | 8.596 ↓ | 9.996 ↓ |

**Reranking Results for Submitted Runs**

Lastly, we simply consider the submitted runs as different summarization models, and rerank the candidate summaries of different submitted runs by our approach. The results on DUC2004 are shown in Table 12. We can see that our reranking algorithm can achieve better ROUGE score than the best run. The results further validate the efficacy of our reranking strategy.

Table 12. Reranking results based on submitted runs

|  | DUC2004 |
|---|---|
| Upper Bound | 11.76 |
| Best run | 9.178 [8.361-10.033] |
| RankSVM | **9.545** [8.823-10.312] |

## Conclusion and Future Work

In this paper, we observe that different summarization models have variable performances over different document sets. Based on this observation, we proposed a two-stage approach for multi-document summarization. In the first stage, we explore an ILP framework to produce candidate summaries for each document set. In the second stage, we propose to use Ranking SVM to rerank the candidate summaries. In this way, the overall summarization performance can be improved. Evaluation results on two DUC datasets verify the efficacy of our proposed approach and the usefulness of the features.

In future work, we will try to make use of more summarization methods to produce more candidate summaries for reranking. We will also investigate advanced deep learning techniques to derive more useful features. For example, we can make use of recurrent neural network or recursive neural network to obtain more reliable semantic representations of sentences, summaries and documents via compositional semantic computation, and then derive new features based on the new semantic representations.